# LungCRCT: Causal Representation based Lung CT Processing for Lung Cancer Treatment

Daeyoung, Kim

*Abstract*— Due to silence in early stages, lung cancer has been one of the most leading causes of mortality in cancer patients world-wide. Moreover, major symptoms of lung cancer are hard to differentiate with other respiratory disease symptoms such as COPD, further leading patients to overlook cancer progression in early stages. Thus, to enhance survival rates in lung cancer, early detection from consistent proactive respiratory system monitoring becomes crucial. One of the most prevalent and effective methods for lung cancer monitoring would be low-dose computed tomography(LDCT) chest scans, which led to remarkable enhancements in lung cancer detection or tumor classification tasks under rapid advancements and applications of computer vision based AI models such as EfficientNet or ResNet in image processing. However, though advanced CNN models under transfer learning or ViT based models led to high performing lung cancer detections, due to its intrinsic limitations in terms of correlation dependence and low interpretability due to complexity, expansions of deep learning models to lung cancer treatment analysis or causal intervention analysis simulations are still limited. Therefore, this research introduced LungCRCT: a latent causal representation learning based lung cancer analysis framework that retrieves causal representations of factors within the physical causal mechanism of lung cancer progression. With the use of advanced graph autoencoder based causal discovery algorithms with distance Correlation disentanglement and entropy-based image reconstruction refinement, LungCRCT not only enables causal intervention analysis for lung cancer treatments, but also leads to robust, yet extremely light downstream models in malignant tumor classification tasks with an AUC score of 93.91%.

## I. INTRODUCTION

Still being one of the most prevalent cancers worldwide, lung cancer is estimated to trigger 226,650 new cases and 124,730 death cases in the United States(US) alone in 2025 [1]. According to the 2020 GLOBOCAN estimates, lung cancer was found to be the leading cause of death among all cancer types by being responsible for 1.8 million death cases worldwide [2]. Though being the second most prevalent cancer after female breast cancer in terms of new cases, lung cancer still accounted for 11.4% of all new cancer cases in 2020 worldwide (2,206,771 new cases). Significance of lung cancer not only derives from high prevalence, but also derives from the silence in early stages. Though it is known that early intervention to lung cancer patients can significantly increase 5-year survival rates regarding prognosis [5], typical lung cancers are known to exhibit no specific symptoms in early stages [6]. Furthermore, major symptoms of lung cancer, such as chest pain, shortness of breath, or consistent coughing [4], are hard to differentiate with other respiratory disease symptoms such as COPD, further leading patients to overlook cancer progression in early stages. Thus, it is crucial for high-risk groups of lung cancer to not only deliberately avoid excessive exposure to risk factors such as smoking, air pollution, or asbestos [3], but also regularly monitor tissue changes within the lung with medical examinations, such as low-dose computed tomography(LDCT) chest scans. Regarding CT scan-based lung cancer detection, computer vision models under deep learning are being actively implemented to enhance conventional tumor detection and classification performance. For example, CNN-based models, such as ResNet50, DenseNet121, InceptionV3, or EfficientNet, Vision Transformer-based models, such as DeiT-Large or Swin-Large, and transfer learning based CNN ensembles exhibited superior performance in classifying CT images into normal, benign, or malignant lung cancer cases with extreme classification accuracies and f1-scores [7, 8, 9, 10].

However, though recent deep learning-based computer vision models succeeded in achieving high lung cancer detection performance, there exist certain limitations in the aspect of model explainability and the possibility of clinical applications. Although XAI methods such as GradCAM can be applied to alleviate the nature of CNNs as a black box and to provide insights regarding what image regions on which the model focused, current approaches do not fully provide clarity in the overall process of diagnosis process itself. Furthermore, as most works only focus on detection performance itself, despite high performance, vision models highly rely on correlations, which can lead the diagnostic model to depend on spurious correlations that are distinct from ground truth causal mechanisms. This correlation-dependent feature limits deep learning model's ability to be applied to clinical treatment simulations or causal intervention analysis for lung cancer alleviations. To address this issue, this research attempted a latent causal representation learning approach for lung cancer analysis. By incorporating a dual objective approach, which (i) aims to extract latent causal factors that can correspond with the underlying physical causal mechanism of lung cancer staging, and (ii) aims to achieve competitive performance in the aspect of downstream lung cancer classification tasks, to Convolutional VAE frameworks with weak guidance from advanced graph autoencoder based causal discovery algorithms and distance Correlation(dCor) measures, it was found that the proposed framework can enhance the applicability of vision models in lung cancer clinical treatments and enable construction of robust, lightweight downstream models for lung cancer detection.

## II. Preliminaries

### A. Related Works

Regarding the use of AI in lung cancer risk detection or classification, significant breakthroughs have been present for the past few years. In [7], based on 2D, 3D lung tumor ROI images acquired from National Lung Screening Trial (NLST) based CT scans, multiple deep CNN-based models, such as ResNet50, DenseNet121, InceptionV3, ShuffleNetv1, or SqueezeNet, and ViT-based models, such as DeiT-Large, BEiT, or Swin-Large, were trained and tested to provide valid risk predictions (malignant or benign) of lung cancer. Regarding best test performance, ResNet50 was found to achieve the highest accuracy of 81%, while MobileNet family succeeded in achieving a superior AUROC score of 86%. In [11], implementations of improved snake optimization (iSOA) to optimize CNN parameters: kernel number, size, and rate of dropout, were attempted under preprocessing steps such as median filtering, gamma correction-based contrast enhancement and SMOTE-based augmentation to achieve high performing lung cancer detection vision models, whereas [10] implemented an AlexNet architecture approach for malignant / non-malignant lung classification tasks, which each led to f1-scores of 91.53% and 96.40% under IQ-OTH/NCCD datasets [12].

Meanwhile, among recent works, [13] attempted a Dual-Branch Modal Classification Approach (DbMCA) to the LIDC-IDRI CT dataset [14] (original image + segmentation mask) to extract both spatial features and high-level patterns that can provide a more holistic diagnosis regarding lung cancer severity, leading to a 91.21% accuracy score and a 91.18% f1-score in 20k sample-based experiments. [8], on the other hand, proposed LECNN, a transfer learning based CNN ensemble specifically designed for lung cancer classification. With pretrained CNN models such as GoogleNet and EfficientNet, majority voting-based LECNN achieved a highest accuracy of 94.98% for raw IQ-OTH/NCCD CT scan data [12], and 99% for augmented IQ-OTH/NCCD dataset.

Though performance of lung cancer detection based on CT images were consistently improved, studies that go beyond detection objectives or attempts to overcome correlation-dependent black box model architectures were found to be few to none. Thus, this research achieves its significance in the aspect of expanding deep learning based lung CT analysis to the area of intervention analysis or clinical treatment simulations, while enabling the retrieval of a more reliable and robust latent space that can correspond with causal factors within the physical mechanism of lung cancer development and progression.

### B. Domain Causal Knowledge: Lung Cancer Progression

Though the specific pathology of lung cancer progression can be diverged based on various tumor sub-types, the fundamental physical causal mechanism of lung cancer development can be represented into a causal graph depicted in Fig 1. First, genetic mutations in genes, such as EGFR, KRAS, MYC and BCL2, trigger cell cycle damage within the lung [15, 16], which leads to carcinogenesis in the lung parenchyma or bronchi. This carcinogenesis process then leads to the initial formation of tumors within the lung parenchyma or bronchi. Then, pro-angiogenic factors, which are mainly represented by vascular endothelial growth factor (VEGF), are activated due to cancerous tumors, leading to angiogenesis which leads to the alleviation of tumor hypoxia [16, 17].

With the continuous growth of cancerous lung tumors based on angiogenesis, invasions to other lung sectors such as visceral pleura, parietal pleura, or even great vessels take place as a result of lung cancer progression [18]. Throughout this process, regional lymph nodes within the lung are affected (involved), leading to enlargements in the hilar, mediastinal, peribronchial or supraclavicular lymph nodes [18, 19]. Thus, when focusing on physical factors visible in CT scans, the fundamental mechanism of lung cancer development or the determination process of lung cancer severity can be summarized into a causal graph in Fig 1. Using this causal graph as the ground truth DAG, this research attempted to retrieve valid latent causal representations of physical causal factors that define lung cancer, leading to the possibility of valid clinical treatment (causal intervention) analysis within lung cancer patients under deep learning-based vision models.

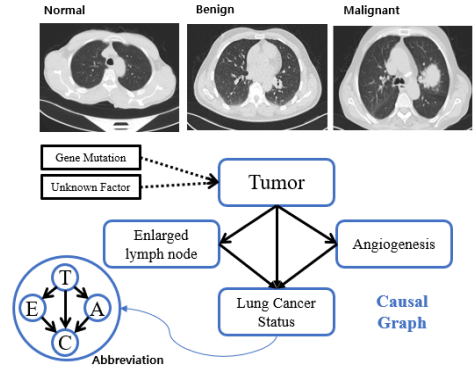

**Fig 1. Causal (Physical) Mechanism of Lung Cancer**
(First row: Lung CT scans of Normal/Benign/Malignant cases)

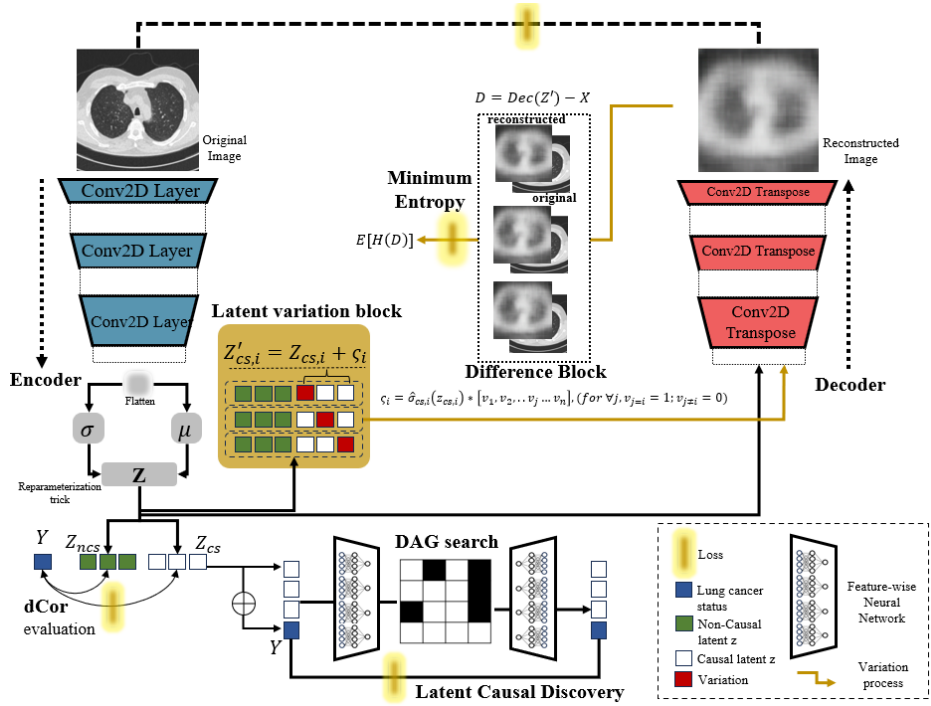

**Fig 2. Visual Summary of the overall LungCRCT framework**
(Latent Causal Encodings: $Z_{cs}$ were used as input for downstream lung cancer classification after LungCRCT training)

## III. METHODOLOGY

### A. Data

This research implemented the IQ-OTH/NCCD CT dataset [12, 20], an open access CT scan public dataset widely used in training lung cancer vision models. Containing a total of 1190 CT scan slices from 110 cases(40: malignant, 15: benign, 55: normal cases) from lung cancer patients and healthy subjects, IQ-OTH/NCCD provides DICOM format lung CT scans from Siemens SOMATOM scanners. With each scan slice representing different angles and sides of the human chest, 120 scan images were classified as benign cases, 561 CT images were classified as malignant, and 416 images were classified as normal. Specific full CT images can be accessed through either Mendeley Data [12] or Kaggle [21]. For analysis, this research resized all CT scans into 256 x 256 format. In terms of training data, to consider label imbalance found in benign cases, 80 images were equally random sampled from each category (normal, benign and malignant, total of 240 CT images). In terms of test data, 40 benign images left after training set definition and randomly sampled 80 images from normal/malignant scans (normal: 40, malignant: 40) were used for lung cancer detection downstream model testing (total of 120 CT scans). Furthermore, for deep learning based encoders to extract only lung-related information, each CT scan was further symmetrically cropped into 180 x 180 formats based on the original center point from 256 x 256 format CT scans. For label information which corresponds to each CT scans, normal, benign, and malignant cases were sequentially re-labelled into numerical values: 0, 1 and 2 to represents stages or severities of lung cancer (*will be denoted as label data $Y$ here for on).

### B. Proposed LungCRCT Framework

The proposed LungCRCT framework (Fig 2.) attempts to retrieve robust causal representations that can fulfill both objectives: (i) clinical applicability of computer vision models in terms of intervention analysis, (ii) reliable robust and low cost construction of downstream lung cancer detection model. By attaching an advanced graph autoencoder-based causal discovery algorithm to the latent space acquired from basic convolutional VAE(CVAE) settings, underlying physical factors and causal links dealt in Fig 1 were extracted to enable intervention analysis in both quantitative and qualitative (visual) aspects. Furthermore, to enhance both the quality of CT scan image generations and the possibility of valid causal space extraction, latent space disentanglements based on feature-wise distance Correlation(dCor) metrics were applied within CVAE encoder training steps. Regarding decoder steps, to reinforce one-to-one matching with physical factors in lung cancer development and enable sensitive qualitative intervention analysis in terms of visual components, a Shannon's Entropy-based variation block which generates CT scan reconstructions under latent causal feature-wise variations was introduced within the overall LungCRCT framework depicted in Fig 2. Specific structures of the LungCRCT framework were constructed as follows.

*(1) Advanced Graph Autoencoder based Causal Representation Learning under Convolutional VAE Structures*

To extract causal representations under dual objectives, latent space was first retrieved from Convolutional VAE (CVAE) processing with an advanced graph autoencoder-based causal discovery algorithm attached which can refine latent features to represent physical factors of lung cancer progression. When defining CT scan images as observational space X and latent space as Z, latent causal structure from X can be defined as in equation (1) to (3). Here, $Z_{cs}$ and $Z_{ncs}$ each denote causal/non-causal latent sub-space defined within Z, which can be acquired by simply splitting total latent space to disjoint sets.

$$Z = f_1(X), \; f_1^{-1}(Z) = X \; (f_1: CVAE \; encoder) \quad (1)$$
$$Z_{cs} = A^T Z_{cs} + \varepsilon, \; \varepsilon: Noise, \; DAG \; \mathcal{G} = \langle A, Z_{cs} \rangle \quad (2)$$
$$Z_{cs} \cup Z_{ncs} = Z, \; Z_{cs} \cap Z_{ncs} = \emptyset \quad (3)$$

To model causal relationships between features within causal latent space $Z_{cs}$, Pearl's Structural Causal Model (SCM) [22,23] was assumed as in equation (2). While being an extension of Structural Equation Model (SEM) to causality, the adjacency matrix $A$ within SCM equation of (2) is assumed to satisfy specific conditions of Direct Acyclic Graphs(DAG) [24]. For the LungCRCT framework, the assumption of (2) regarding latent causal variables $Z_{cs}$ was expanded to equations (4) and (5), based on the approach of Graph Autoencoder-based causal discovery algorithm [25], which expands NOTEARS [26] beyond linearity. By incorporating non-linearity to causal modeling with a non-linear encoder-decoder structures as in (4) and implementing the idea of message passing operation with $A^T$, Graph Autoencoder based causal discovery algorithm was found to succeed in retrieving true causal graphs when compared to existing models such as DAG-GNN [27] and NOTEARS in previous works [25,26]. Regarding adjacency matrix $A$, components in $A$ were considered as trainable weights under continuous gradient descent algorithms.

$$Z = f_1(A^T f_2(Z)) + \varepsilon, \; f_{1(2)}: nonlinear \; Encoder(Decoder) \quad (4)$$
$$H(A) = tr(e^{A \odot A}) - d, \; A \in \mathbb{R}^{d \times d}, Z \in \mathbb{R}^{N \times d} \quad (5)$$

For a more flexible approach, this research improved the application of Graph Autoencoder based causal discovery algorithms by incorporating two additional conditions: (i) a feature-wise non-linear function assumption and (ii) an alternative DAG constraint definition introduced in DAGMA [28]. First, for the original encoder-decoder structure, separate neural networks were assumed to approximate $f_1$ and $f_2$ under a generalized additive model approach, leading to an alternative causal model definition of (6). Second, regarding the original DAG constraint: (5), the log-determinant based constraint of (7) was implemented to replace (5) due to its advantages in better cycle consideration and gradient behavior [28]. Specifically, if the weighted adjacency matrix $A$ has a cycle of length $k$, the contribution of the existing cycle is diminished by $1/k!$ under the DAG constraint definition of (5). On the other hand, due to log-det function being bounded away from 0, detection of larger cycles becomes convenient under definition of (7). Furthermore, though having identical computational complexity of $\mathcal{O}(d^3)$, it was found that the log-det based DAG constraint works faster in returning constraint values in practice [28].

$$for \; \forall i, \; \widehat{Z}_{cs,i} \simeq f_{2,i}(A_i^T f_{1,i}(Z)), f_{1(2),i}: i^{th} \; NN \quad (6)$$
$$H_{ldet}^s(A) = -\log\{\det(sI - A \odot A)\} + d \log(s), s > 0 \quad (7)$$

Based on these settings and an augmented lagrangian approach for optimization [25,26] latent causal discovery loss function within the LungCRCT framework was defined as (8) and (9). Here, instead of sole use of $Z_{cs}^{(i)}$, lung cancer status label: Y was concatenated to $Z_{cs}^{(i)}$ to provide weak supervision within causal discovery steps. $\widehat{Z_{cs}}^{(i)}$ in (8) is the direct result of application of equation (6) to latent causal representations: $Z_{cs}^{(i)}$ from CVAE lung CT encodings.

$$GAE \; input: Q_{cs} = \langle Z_{cs}, Y \rangle$$
$$\min_{A,\theta_1,\theta_2} \mathcal{L}_2 = \min_{A,\theta_1,\theta_2} \frac{1}{n} \sum_{i=1}^n \left\| Q_{cs}^{(i)} - \widehat{Q}_{cs}^{(i)} \right\|^2 + \lambda \|A\|_1 + \alpha h(A)$$
$$+ \frac{\rho}{2} |h(A)|^2; \; h(A) = -\log\{\det(sI - A \odot A)\} + d \log s \quad (8)$$

$$\alpha^{(k+1)} = \alpha^{(k)} + \rho^{(k)} h(A^{(k+1)})$$
$$\rho^{(k+1)} = \begin{cases} \beta \rho^{(k)}, (if \; |h(A^{(k+1)})| \geq \gamma |h(A^{(k)})|) \\ \rho^{(k)}, (o.w.), (\beta > 1, \gamma < 1) \end{cases} \quad (9)$$

Meanwhile, for LungCRCT CVAE encodings and image reconstruction tasks, a multiple convolutional layer-based vanilla Variational Autoencoder (VAE) with a latent space bottleneck was assumed. That is, observational variables: X from lung CT scans were compressed into low-dimensional latent space Z to extract only informative features. Similar to tabular VAE loss settings, the Evidence Lower Bound (ELBO) approach was also applied for CVAE loss definition [29]. Regarding distributional assumptions for log likelihood computations, Bernoulli distribution was implemented, for all image data pixels were normalized to values within 0 to 1. With additional assumptions of $q_\phi(z|x) \sim N(z, I)$ and $p(z) \sim N(0, I)$, original ELBO loss of (10) was further reduced to the form of (11) (binary cross entropy + regularization term). Here, $v$ is an additional parameter which was assumed in practical aspects to prevent collapse or convergence to limited images in CVAE image reconstruction tasks.

$$-\log(p(x)) = -\int \log\left(\frac{p(x,z)}{q_\phi(z|x)}\right) q_\phi(z|x) dz + KL(q_\phi(z|x) \| p(z|x))$$
$$\geq -\int \log\left(\frac{p(x|z)p(z)}{q_\phi(z|x)}\right) q_\phi(z|x) dz = -\text{ELBO}(\phi)$$
$$= -E_{q_\phi}(\log(p(x|z))) + KL(q_\phi(z|x) \| p(z))$$
$$\simeq -\log \prod_{j=1}^D p(x^{(j)}|z^l) + KL(q_\phi(z|x) \| p(z)), (L = 1) \quad (10)$$

$$CVAE \; loss: \mathcal{L}_1 = \sum_j^D \log p_{ij}^{x_{ij}} (1 - p_{ij})^{1-x_{ij}} + \frac{1}{2} v \sum z_{ij}^2 \quad (11)$$

*(2) dCor-based Latent Space Disentanglement*

For latent space disentanglement, distance correlation(dCor) [30] was implemented as an alternative to Mutual Information (MI). Though MI is well known for its superiority in detecting feature-wise relevance in both linear/non-linear aspects, implementing MI within neural network training creates high burdens in terms of cost due to its nature which requires probability density function estimations of all target variables. Thus, within the LungCRCT framework, the dCor metric was implemented as an alternative to evaluate informativeness of features in latent space Z regarding label data Y to enhance the possibility of splitting causal information into $Z_{cs}$ and non-causal information, such as CT angles, posture, or initial organ size, into $Z_{ncs}$.

Distance Correlation (dCor) is a dependency measure introduced by G.J, Szekely et al.(2007) which computes the degree of joint independence between random vectors with arbitrary dimensions under Euclidean distances from sample elements, making it possible to detect non-linear dependencies unlike most conventional correlation measures such as Pearson's correlation coefficient [30, 31]. Specific definitions and derivation process of the empirical dCor can be summarized as equation (12) and (13).

$$dCor: \mathcal{R}(X,Y), \mathcal{R}^2(X,Y) = \begin{cases} \frac{\mathcal{V}^2(X,Y)}{\sqrt{\mathcal{V}^2(X)\mathcal{V}^2(Y)}}, & (\mathcal{V}^2(X)\mathcal{V}^2(Y) > 0) \\ 0, & (\mathcal{V}^2(X)\mathcal{V}^2(Y) = 0) \end{cases} \quad (12)$$

$$s.t. \; \mathcal{V}^2(X) = \frac{1}{n^2}\sum_{k,l}^{n} A^2_{kl}, \mathcal{V}^2(Y) = \frac{1}{n^2}\sum_{k,l}^{n} B^2_{kl}, \mathcal{V}^2(X,Y) = \frac{1}{n^2}\sum_{k,l}^{n} A_{kl}B_{kl}$$

For random sample $(X,Y) = \{(X_k, Y_k): k = 1,2,...n\}, X \in \mathbb{R}^p, Y \in \mathbb{R}^q$
$A_{kl} = a_{kl} - \bar{a}_{k.} - \bar{a}_{.l} + \bar{a}_{..}, B_{kl} = b_{kl} - \bar{b}_{k.} - \bar{b}_{.l} + \bar{b}_{..}$
$a_{kl} = |X_k - X_l|_p, \bar{a}_{k.} = \frac{1}{n}\sum_{l=1}^{n} a_{kl}, \bar{a}_{.l} = \frac{1}{n}\sum_{k=1}^{n} a_{kl}, \bar{a}_{..} = \frac{1}{n^2}\sum_{k,l=1}^{n} a_{kl}$ **(13)**

Due to its definition, dCor values are computed within the range of 0 to 1, making it easier to intuitively evaluate the strength of variable dependencies within data. In the LungCRCT framework, dCor was applied to the latent space Z from CVAE encodings. By minimizing feature-wise dCor values between non-causal latent sub space: $Z_{ncs}$ and lung cancer status $Y$, and maximizing feature-wise dCor values from $Z_{cs}$ and $Y$, disentanglement of lung cancer related/non-related visual information was attempted. Furthermore, within training, to emphasize the importance of $Z_{cs}$, an additional weighting parameter $\tau$ was incorporated to the dCor loss function definition described in equation (14).

$$dCor\;Loss: \mathcal{L}_3 = \min_{Z} \frac{1}{K}\sum_{k=1}^{K} dCor(Z_{ncs,k}, Y) - \tau \frac{1}{J}\sum_{j=1}^{J} dCor(Z_{cs,j}, Y) \quad (14)$$

*(3) Entropy-based Latent Variation Block Design*

To further enhance the possibility of valid clinical causal intervention analysis using LungCRCT, an entropy-based latent variation block that enhances sensitivity of the CVAE decoder was further assumed. For causal information retrieved from previous structures to be expanded to causal intervention simulations, valid visual reconstructions of CT scans, which can accurately represent changes within the patient's lung when a certain physical factor within the lung cancer mechanism is intervened, should be able to be generated with significance. That is, sensitive reaction of the decoder to variations in causal representation values becomes a prerequisite to enable real world clinical treatment analysis. The LungCRCT framework directly incorporates this prerequisite with a latent variation block which first creates causal feature-wise variations, and then evaluates change generated in corresponding reconstructed images under Shannon's entropy. Based on the approach suggested in [32], LungCRCT evaluated randomness of images regarding difference between generated images(before/after latent causal representation variation) with Shannon's entropy estimation. If visual transformations from varying latent causal factor z does not match with changes in corresponding tissues or lung components, visualized difference from reconstructions would be close to random (maximum entropy), whereas if visual transformations from varying z match with changes in corresponding tissue or lung components, estimated entropy values would be far less than maximum entropy values (ex: Fig 3). Based on this assumption, specific mechanisms and loss for the latent variation block were constructed as (15) and (16).

$$i^{th}\;latent\;causal\;factor\;variation: Z_{cs,i}' = Z_{cs,i} + \varsigma_i$$
$$s.t. \; \varsigma_i = \hat{\sigma}_{cs,i}(Z_{cs,i}) * \langle v_1, v_2, ..v_j ... v_d\rangle, (for\;\forall j, v_{j=i} = 1, v_{j\neq i} = 0) \quad (15)$$

$$Variation\;loss\;\mathcal{L}_4 = E[H(D)] \approx \frac{1}{d}\sum_{i}^{d} H(D_i),$$
$$where\;D_i = \{G(Z_i) - G(Z_i')\}^2, G: CVAE\;decoder$$
$$Z_i' = \langle Z_{ncs}, Z_{cs,i}'\rangle, H(p) = -\sum p_i \log p_i \quad (16)$$

Here, regarding variation: $\varsigma_i$, 1-sigma impulse was assumed when altering causal feature: $Z_{cs,i}$. For probability estimations in computing entropy $H$, histograms with equal width interval were generated from data: $D_i$ in (16), with number of bins set as a tunable hyperparameter.

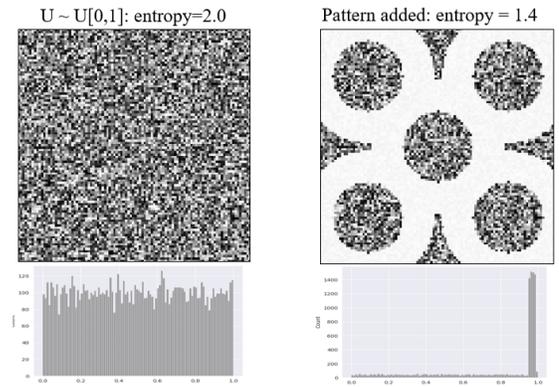

**Fig 3. (Left)** Uniform noise based image (Entropy=2.0) **(Right)** Circle patterns added in image (Entropy=1.4).

*(4) LungCRCT framework Total Loss*

Based on previously proposed sub-structures within the LungCRCT framework, total loss of LungCRCT was defined as (17). Here, four additional weight parameters: $\varphi_1$ to $\varphi_4$, were applied to achieve balance among sub-objectives within stages of training. In early stages, CT image reconstruction was most emphasized for initial latent space refinements. In mediate stages, both $\mathcal{L}_2$ and $\mathcal{L}_3$ were emphasized for latent space optimization. In final stages, $\mathcal{L}_4$ was additionally significantly considered for the CVAE decoder refinement. Based on this approach, neural network weights were trained with high flexibility under initial objectives: (i) clinical applicability of computer vision models in terms of lung cancer causal intervention analysis and (ii) reliable low cost construction of downstream lung cancer detection model.

$$\mathcal{L}_{total} = \varphi_1 \mathcal{L}_1 + \varphi_2 \mathcal{L}_2 + \varphi_3 \mathcal{L}_3 + \varphi_4 \mathcal{L}_4$$
Loss $\mathcal{L}_1$: CVAE, $\mathcal{L}_2$: causal, $\mathcal{L}_3$: disentanglement, $\mathcal{L}_4$: variation **(17)**

### C. Experimental Design

Under the proposed LungCRCT framework, 240 chest CT images from the IQ-OTH/NCCD dataset and lung cancer status labels: $Y$ were implemented as input training data. For entropy loss computations in latent variation blocks, 50 random images from train data were used to limit excessive computational cost. Regarding Convolutional VAE reparameterization, standard gaussian distribution was assumed to create randomness. Specific substructures and hyperparameters within the LungCRCT framework were constructed as TABLE I, TABLE II. This research assumed five consecutive Convolutional2D layers for the CVAE encoder, with the number of filters in each layer set as 16, 32, 32, 16 and 16. For the CVAE decoder, all Conv2DTranspose layers except for the last hidden layer were set symmetrically with the proposed CVAE encoder. Regarding the last hidden layer, 32 filters were implemented instead of 16 filters to enhance expressiveness in visual reconstruction procedures.

Meanwhile, regarding the dimension of latent space $Z$, 5 non-causal latent features and 3 causal latent features were assumed. Here, the number of dimensions for features in $Z_{cs}$ was selected according to the number of physical lung cancer related factors (tumor, angiogenesis, lymph node enlargement) dealt in *II. Preliminaries*. Therefore, the input dimension of the advanced GAE was defined as 4(=3+1) due to lung cancer status label variable $Y$ being concatenated for weak guidance.

To directly link weighted adjacency matrix $A$ with GAE encoder-decoder structures for integrated weight training, additional adjacency weight layer was defined in front of the original decoder of GAE. For the adjacency layer, zero-initialization for layer weights was implemented to prevent bias deduced from specific initial starting points in DAG structure search. Moreover, when defining the adjacency layer, self-loops and links from lung cancer status($Y$) to causal features in $Z_{cs}$ were blacklisted for efficient causal discovery.

TABLE I
NETWORK SETTINGS FOR CVAE

| Type | Network Architecture |
|---|---|
| CVAE Encoder | 16 filters, Conv2D 6x6, stride: 2, silu<br>32 filters, Conv2D 6x6, stride: 2, silu<br>32 filters, Conv2D 4x4, stride: 2, silu<br>16 filters, Conv2D 4x4, stride: 2, silu<br>16 filters, Conv2D 3x3, stride: 2, silu<br>128, FC layer(Flattened), ELU<br>16, FC layer, ELU<br>16(=(5+3)*2), FC layer, Linear |
| CVAE Decoder | 16, FC layer, ELU<br>128, FC layer(Flattened), ELU<br>4x4x16 FC layer, ELU<br>Reshape(4,4,16) layer<br>16 filters, Conv2DTranspose 3x3, stride: 2,silu<br>32 filters, Conv2DTranspose 4x4, stride: 2,silu<br>32 filters, Conv2DTranspose 4x4, stride: 2,silu<br>32 filters. Conv2DTranspose 6x6, stride: 2,silu<br>3 filters, Conv2DTranspose 6x6, stride: 2,sigmoid |

*Silu: Swish activation function from [33].

TABLE II
NETWORK SETTINGS FOR GAE

| Type | Parameter settings |
|---|---|
| $f_1$ (GAE Encoder) | 16(=4*4)(# of nodes), Sparse Layer, ELU<br>16(=4*4), Sparse Layer, ELU<br>4, Sparse Layer, Linear |
| $f_2$ (GAE Decoder) | **4, Adjacency matrix layer, Linear**<br>16(=4*4), Sparse Layer, ELU<br>16(=4*4), Sparse Layer, ELU<br>4, Sparse Layer, Linear |

*Sparse Layer: Masked FC Layer based on additive modeling approach

For feature-wise NN training regarding the GAE encoder-decoder, binary masks were Hadamard multiplied to network weights to ensure the generalized additive approach. For each latent feature, separate NNs with two hidden layers (4 nodes layer-wise) and an Elu activation function was assigned. Masking fully connected layers to create feature-wise NNs was proceeded through transforming binarized mask matrices into neural network kernel constraints with tensorflow.keras.constraints.Constraint(). To construct overall model sub-structures and implement training under LungCRCT frameworks, Python Tensorflow (version 2.19.0) was used. All experiments from the proposed LungCRCT framework were proceeded with Google Colab's basic T4 NVIDIA GPU environments. Overall training was based on **Algorithm 1** and full-batch gradient descent settings.

---

**Algorithm 1** Training under LungCRCT framework

-**Total number of epoch(K) = 450; i =0;**
-**Initial Settings:** $\alpha = 0.6, \rho = 0.1, \gamma = 0.9, \beta = 1.01$, Adam Optimizer
-**While i < K**:
   1. Compute $\mathcal{L}_1, \mathcal{L}_2, \mathcal{L}_3, \mathcal{L}_4$; $\tau = 1.5, v = 0.001, \lambda = 0, s = 1$
   2. Compute Total LungCRCT loss: $\mathcal{L}_{total}$
   **IF i < 50**:
     $\mathcal{L}_{total} = \mathcal{L}_{total}(\varphi_1 = 1, \varphi_2 = 0.5, \varphi_3 = 0, \varphi_4 = 0)$
   **ELIF i < 100**:
     $\mathcal{L}_{total} = \mathcal{L}_{total}(\varphi_1 = 1, \varphi_2 = 1, \varphi_3 = 1, \varphi_4 = 0)$

```
        ELSE:
            𝓛_total = 𝓛_total(φ₁ = 1, φ₂ = 1, φ₃ = 1, φ₄ = 1)
        3. Update weights via gradient descent algorithm:
            1) Update GAE weights and adjacency matrix A with 𝓛₂
                *(learning rates: 0.007 (A), 0.002 (GAE))
            2) Update CVAE structure weights with 𝓛_total
                *(learning rates: 0.001(enc), 0.0007(dec))
            3) Update α, ρ using (9)
        5. Update i = i+1;
        Return: Latent features Z, fitted adjacency matrix A
```

When training under the LungCRCT framework was completed, validity of latent causal representations from the CVAE encoder and causal adjacency matrix was further checked. In quantitative aspects, Structural Hamming Distance(SHD) from the extracted weighted adjacency matrix and true causal adjacency matrix from the causal graph illustrated in Fig 1., was compared to evaluate success in causality detection or retrieval. When computing SHD from LungCRCT training, fitted (weighted) adjacency matrix was binarized by the criterion of converting top 30% weights in terms of absolute values to 1 (causality) and others as 0 (no causality). Comparison between causal graphs was based on best possible match in topological aspects. In qualitative aspects, individual change in visual reconstructions generated from varying individual latent causal variables with others being fixed was checked whether it matches characteristics of ground truth causal factors within the lung cancer progression mechanism.

### D. Downstream Task Expansion: Lung Cancer Classification

With the success acquired from validating causal representation learning results from LungCRCT frameworks, $Z_{cs}$ based downstream lung cancer detection model was trained and tested within final stages of analysis. That is, only latent causal representations from LungCRCT were used to train lung cancer detection models under the assumption that features in sub-space $Z_{cs}$ are sufficient statistics of lung cancer-related visual data. Regarding the downstream model structure, a simple DNN with model architectures of TABLE III was used for fitting. Here, to create a downstream model that can classify malignant/non-malignant lung cases, the original label data Y was converted to a binary label of 0 and 1 by setting malignant cases as 1 and normal/benign cases as 0.

For downstream model training, identical train data implemented in LungCRCT fitting was used, whereas for the evaluation of fitted downstream model, test dataset defined in *III. Methodology* Section. A. was implemented. Regarding hyperparameter settings, total number of epochs, type of optimizer, learning rate and batch size were set as 300, Adam optimizer, 1e-4 and 80. For evaluation metrics, classification accuracy, Macro f1-score, Macro precision, Macro recall and AUC scores from test data processing were implemented as performance measures. Based on these settings, validity and effectiveness of LungCRCT frameworks in both objectives (i) and (ii) were thoroughly checked.

TABLE III
HYPER PARAMETER SETTINGS IN DNN

| Type | Network Structure |
|---|---|
| DNN | Batch Normalization( ) <br> 32(# of nodes), FC layer, ELU <br> 32, FC layer, ELU <br> 1, FC layer, Sigmoid |

*Loss Settings: Binary Cross entropy*

## IV. EXPERIMENTAL RESULTS

### A. LungCRCT Representation Learning Results

Results from training under the LungCRCT framework were deduced as follows. In both total LungCRCT loss and causal discovery loss, successful convergence was found to be achieved from 450 epochs of training. Regarding individual loss values, $𝓛_1$=0.6014, $𝓛_2$=0.0059, $𝓛_3$= -1.4467, $𝓛_4$=0.3003 were deduced. Specifically, $𝓛_4$ was found to be significantly lower than maximum entropy with number of bins set as 5 (max-entropy = $log_2|\mathcal{X}| \approx 2.322$), which implies successful decoder refinement in terms of reconstructions from latent causal representation variations. Thus, success in latent space retrieval, decoder refinement and image generation tasks was found to be achieved under the proposed LungCRCT framework. Regarding lung cancer causal information disentanglements under distance Correlation(dCor) measures, latent feature-wise Mutual Information(MI) was quantified by sklearn.feature_selection package function: mutual_info_regression(n_neighbors=5) to evaluate disentanglement success. Individual MIs for latent features were sequentially computed as follows: 0, 0, 0, 0.0165, 0.0259, 1.1018, 1.1018, 1.1018. With mean MI from $Z_{ncs}$ being 0.008 and mean MI from $Z_{cs}$ being 1.1018, strong disentanglement was checked regarding lung cancer-related/non-related visual factors in CT.

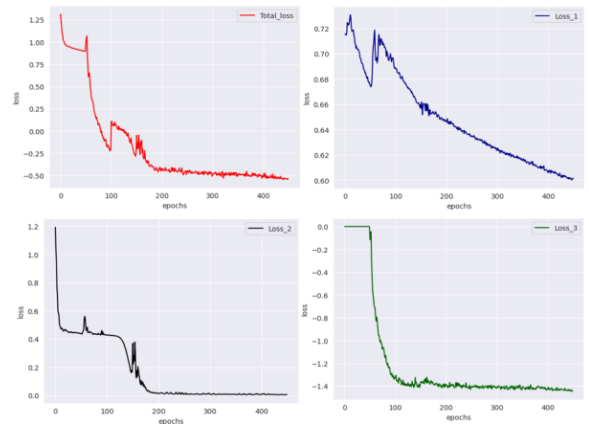

**Fig 4. LungCRCT Loss values. (Upper Left) Total loss, (Upper Right) $𝓛_1$, (Lower Left) $𝓛_2$, (Lower Right) $𝓛_3$**

Furthermore, disentanglements of causal/non-causal CT information were further evaluated in visual aspects using t-

SNE. Under the TSNE(perplexity=20,learning_rate=100) function implemented from sklearn.manifold package, t-SNE planes from train data were visualized as in Fig. 5 with label data Y set as target status. Results show that clear clusters that strongly correspond with each label were found to be generated with no significant overlaps in $Z_{cs}$, leading to the conclusion that dCor-based lung cancer information disentanglement was successfully computed.

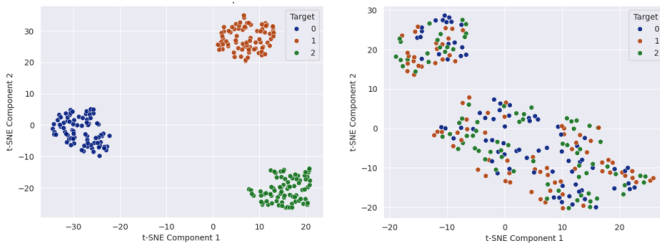

**Fig 5. (Left)** t-SNE result for $Z_{cs}$, **(Right)** t-SNE result for $Z_{ncs}$

Based on these findings, validity of causal retrieval in the extracted latent causal space $Z_{cs}$ was also checked to evaluate success in causal representation extraction. When binarized by the threshold of top 30% absolute weights, the extracted weighted adjacency matrix from LungCRCT and the true causal graph from Fig 1. were found to exhibit Structural Hamming Distance(SHD) of 2.0 (Fig 6) under best possible match (all edges except $Z_{cs,2}$ to $Z_{cs,0}$ satisfy DAG constraints). Thus, considering $\mathcal{L}_2$ convergence, dCor lung cancer related/non-related information disentanglements and a significantly small SHD of 2.0 in extracted latent causal graph structures, it was found plausible to conclude that true latent causal space retrieval from the LungCRCT framework was successfully achieved. Moreover, with high topological similarities between the extracted latent causal graph and true causal DAG found in Fig 6., high possibility of one-to-one correspondence with tumor, lymph node enlargement and angiogenesis was found regarding $Z_{cs,0}, Z_{cs,1}, Z_{cs,2}$, sequentially.

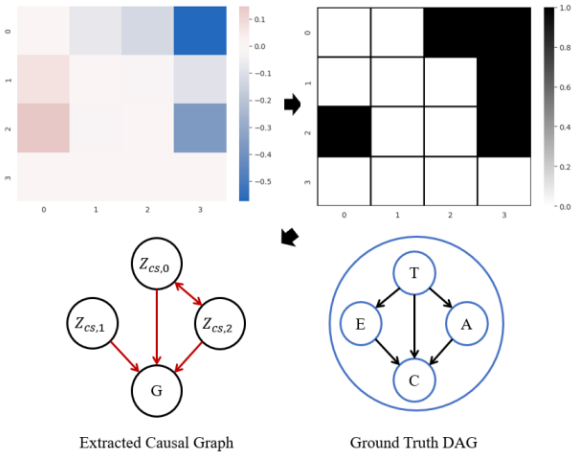

**Fig 6.** Visualization of extracted latent causal adjacency matrix. Upper Left: Weighted, Upper Right: Binarized.

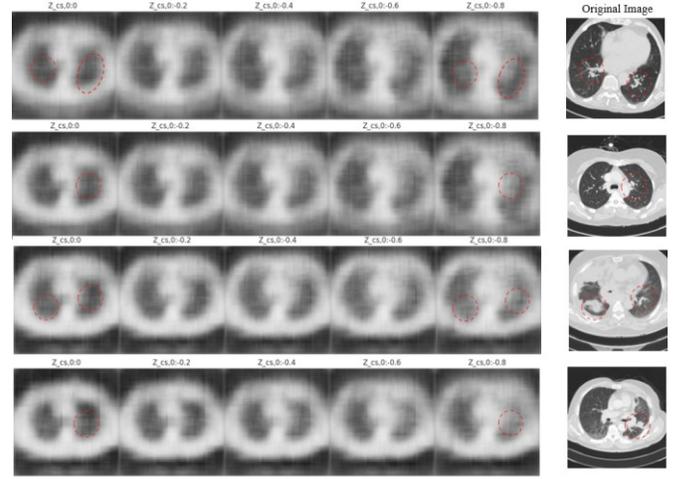

**Fig 7.** Visualizations from varying latent causal feature: $Z_{cs,0}$. Original Image in 6[th] column. Tumorous regions emphasized in red circles.

### B. Qualitative Causal disentanglement checks

Based on success found in LungCRCT-based latent causal space in terms of causal loss convergence and a significantly low SHD, qualitative causal disentanglement checks regarding physical causal factors within the lung cancer progression mechanism were computed based on varying individual $Z_{cs,i}$. Specifically, one-to-one correspondence between factors in $Z_{cs}$ and tumor, angiogenesis and lymph node enlargements was checked by visualizing Lung CT image generations from varying each latent causal representations with others being strictly fixed. When setting variation intervals for image generation, min-max values from $Z_{cs}$ and equal-width interval generation were considered for valid variation analysis. Visualization results from individual feature variation were deduced as Fig 7 to Fig 9.

For variables: $Z_{cs,0}$, when values were decreased from 0.0 to -0.8, regions where tumors are highly suspected within original images were found to be altered (Fig 7). Specifically, as $Z_{cs,0}$ decreased from 0.0 to -0.8, bright hyperdense mass was found to be consistently generated in all examples with regional correspondence. Considering visual characteristics of tumor within CT scans and correspondence in regions, it was found plausible to conclude that $Z_{cs,0}$ not only incorporates valid lung cancer-related causal information, but also has high possibility of having one-to-one correspondence with tumor dealt within the lung cancer progression mechanism introduced in Fig 1.

For variables: $Z_{cs,1}$, when values were decreased from 0.1 to -0.7, hilar and mediastinal lymph node regions were found to be altered (Fig 8). Although changes when varying $Z_{cs,1}$ were found to be not as significant as $Z_{cs,0}, Z_{cs,2}$ variation results, as values were decreased from 0.1, CT scan reconstructions regarding central lymph node regions became less dense, and less enlarged. Regarding regional correspondence and visual characteristics found when varying $Z_{cs,1}$, it was found that $Z_{cs,1}$ can one-to-one correspond with lymph node enlargements within the lung cancer progression mechanism.

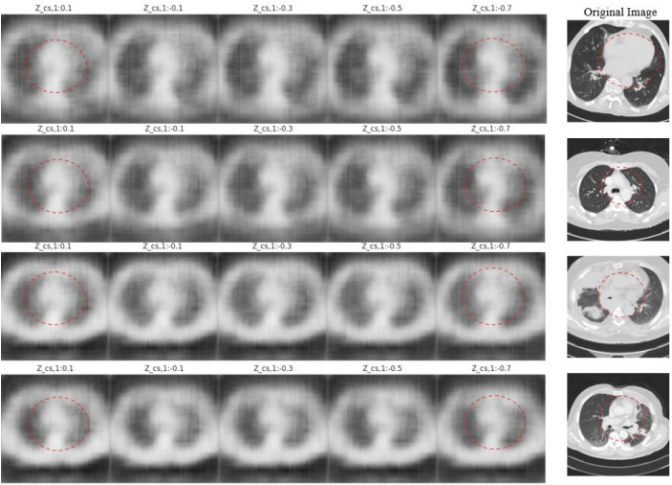

**Fig 8. Visualizations from varying latent causal feature:** $Z_{cs,1}$. Original Image in 6th column. Hilar and Mediastinal lymph node regions emphasized in red circles.

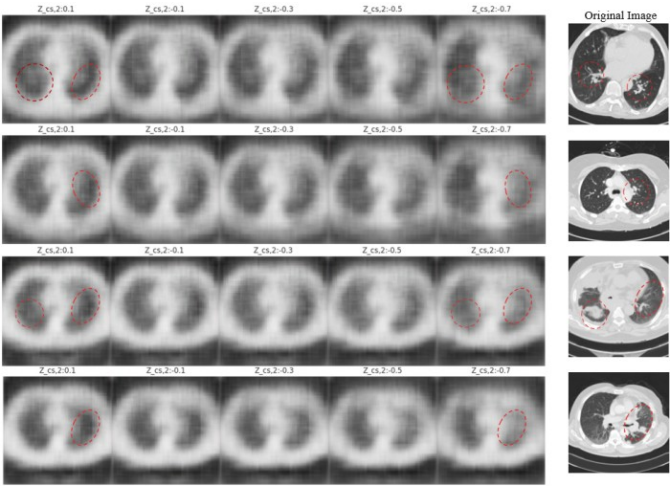

**Fig 9. Visualizations from varying latent causal feature:** $Z_{cs,2}$. Original Image in 6th column. Regions where angiogenesis is suspected, especially regions where tumors, TIB and sunburn signs are suspected were emphasized in red circles.

Meanwhile, regarding latent causal factor $Z_{cs,2}$, it was found valid to conclude that latent representation $Z_{cs,2}$ can have one-to-one correspondence with angiogenesis. When $Z_{cs,2}$ values were decreased from 0.1 to -0.7, hyperdense blurring was found to be consistently generated around tumorous regions, possible feeding vessel sign regions, tree-in-bud regions and sunburst sign regions [34] within original images, while other visual features were fixed. Here, though tumor-suspected regions were found to be altered, specific mass generations were not visible unlike in $Z_{cs,0}$ variation cases (ex: comparison between first rows in Fig 7 and Fig 9. Identical original image implemented), which implies that $Z_{cs,2}$ represents distinct information other than tumor. Considering absence of mass generation, regional correspondence with tumor suspected regions, and the fact that angiogenesis appears as micro vessels in peripheral regions of tumor related regions [35], it was found valid to conclude that $Z_{cs,2}$ may have high possibility of having one-to-one correspondence with angiogenesis within the physical lung cancer mechanism dealt in Fig 1.

With success in causal representation learning exhibited from previous evaluations, LungCRCT can not only return non-linear causal relationships between physical factors in numerical aspects, but also enable intervention analysis or simulations in the aspect of treatments. Based on fitted Graph Autoencoder within the LungCRCT framework and the adjacency matrix regarding latent causal factors, one can get interventional results by first varying a certain physical factor within $Z_{cs}$ and then passing the new input vector into the fitted GAE encoder-decoder structure, similar to the structural causal layer mechanism introduced in [36]. By further passing the newly generated (intervened) latent representations to the CVAE decoder, visual changes in lungs or adjacent lymph nodes due to intervention can be illustrated with validity.

*C. LungCRCT Causal Encoding based Downstream Models: Lung Cancer Malignancy Classification*

With success achieved in latent causal representation learning regarding the physical lung cancer mechanism, the LungCRCT framework was further extracted to conventional downstream tasks. Only using three latent causal representations: $Z_{cs,0}$ to $Z_{cs,2}$, a 2-hidden layer DNN was fitted to classify CT scans into non-malignant or malignant lung cancer cases. Results regarding training and evaluation under train/test datasets were deduced as TABLE IV and Fig 10. Specifically, the LungCRCT-based DNN exhibited 91.67% classification accuracy, 91.88% macro recall score, 90.07% macro precision score, 90.84% Macro F1-score and 93.91% ROC AUC scores. Furthermore, regarding additional metrics, sensitivity of 92.5%, specificity of 91.25% and NPV of 96.05% were deduced from test data classification tasks.

Considering the fact that only 240 lung CT scan images with no augmentation were implemented as train data, and the fact that only three causal representations (encodings) from LungCRCT were used for DNN fitting, deduced test performance from the proposed downstream model can be considered highly significant and competitive in classifying lungs into malignant or non-malignant cases.

TABLE IV
LUNG CANCER CLASSIFICATION PERFORMANCE (LUNGCRCT)

| Data Type | Accuracy | Macro Recall | Macro Precision | Macro F1 | AUC score |
|---|---|---|---|---|---|
| Train Data | 100% | 100% | 100% | 100% | 100% |
| **Test Data** | **91.67%** | **91.88%** | **90.07%** | **90.84%** | **93.91%** |

*Test data classification performance in bold*

Thus, it was found that LungCRCT not only enables causal intervention analysis for lung cancer treatments in visual aspects, but also provides significant starting points regarding causal representation-based lightweight downstream lung cancer classification model constructions with high validity.

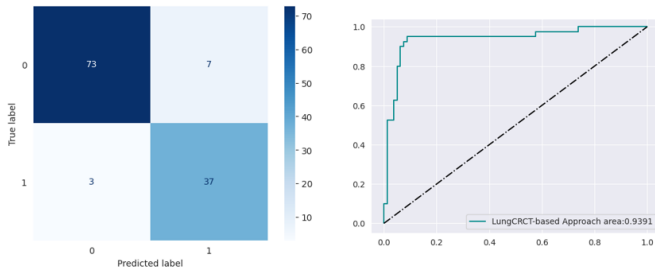

**Fig 10. Lung CT Classification Performance (Malignant/non-Malignant) from LungCRCT framework. (Left) Confusion matrix from test data, (Right) ROC curve from test data**

## V. Conclusion

This study attempts to construct a novel lung CT scan-based latent causal representation learning framework (LungCRCT), which can significantly expand the use of conventional computer vision models in lung cancer analysis from lung cancer status classification tasks to causal intervention analysis and visual treatment simulations. Under the dual objective of (i) and (ii) dealt in *III. Methodology* Section B., a Convolutional VAE with a neural causal discovery algorithm attached to the latent subspace was implemented. Within the LungCRCT CVAE training framework, VAE's latent space was first disentangled based on relevance regarding visual features of lung cancer status, and then was attached to a non-linear neural causal discovery algorithm to extract valid latent causal representations of physical factors from lung cancer progression with only gradient descent approaches.

Specifically, using dCor-based latent space disentanglements, an advanced Graph Autoencoder based causal structure learning, and an entropy-based latent variation block that enhances sensitivity of the CVAE decoder, successful latent causal space retrieval regarding physical factors: tumor, angiogenesis and lymph node enlargements, and success in valid image reconstructions (generations) of chest CT scans were found to be achieved. Thus, trained LungCRCT within this work was found to not only return possible one-to-one causal representations of physical factors found in lung cancer progression, but also enable one to carry out effective clinical intervention analysis under the visualization approach described in IV.B.

Furthermore, usage of causal encodings from LungCRCT was expanded to conventional lung cancer classification tasks to check competitiveness of low-dimensional causal representational space usage in downstream tasks. Though using only a total of 83,613 parameters within the representation encoding + classification process, the proposed downstream DNN model succeeded in returning competitive classification performance when detecting malignant/non-malignant lung CT scans, and even exhibited significant AUC scores of 93.91% and NPV score of 96.05%. Thus, under the use of LungCRCT framework in lung cancer analysis, causal deep learning-based vision models are expected to return both valid lung cancer status of patients and patient-wise intervention simulation results with high significance.

Though immense success has been achieved from the proposed LungCRCT framework, several limitations regarding image clarity and consideration of non-physical factors of lung cancer still remain as a residing issue. Due to the intrinsic nature of VAE reconstructions from latent space, vagueness was consistently found within lung CT scan image reconstructions (or generations). Furthermore, as LungCRCT only relies on single-modal data usage (image data), non-physical risk factors, such as exposure to air pollution or smoking status of patients were not considered within the extracted lung cancer latent causal graph. Thus, for more extensive applications of LungCRCT in clinical fields, it is recommended that these limitations being dealt through measures such as diffusion models or multi-modal patient data.

## VI. Reference

[1] R.L, Siegel et al., "Cancer statistics, 2025*", CA: A Cancer Journal for Clinicians*, Vol. 75, no.1, pp. 10-45, 2025.
[2] H, Sung et al., "Global Cancer Statistics 2020: GLOBOCAN Estimates of Incidence and Mortality Worldwide for 36 Cancers in 185 Countries", *CA: A Cancer Journal for Clinicians,* Vol. 71, no. 3, pp. 209-249, 2021.
[3] C.T, Jani et al., "Evolving trends in lung cancer risk factors in the ten most populous countries: an analysis of data from the 2019 Global Burden of Disease Study", *eClinicalMedicine*, Vol. 79, 103033, 2025.
[4] Centers for Disease Control and Prevention(CDC), "Lung Cancer", 2024, https://www.cdc.gov/lung-cancer/index.html.
[5] F.M, Habbab et al., "Early Detection of Lung Cancer: A Review of Innovative Milestones and Techniques", *Journal of Clinical Medicine*, Vol. 14, no. 21, 7812, 2025.
[6] C, Read et al., "Early Lung Cancer: screening and detection", *Primary Care Respiratory Journal,* Vol. 15, no. 6, pp. 332-336, 2006.
[7] Y, Jiang et al., "A benchmark of deep learning approaches to predict lung cancer risk using national lung screening trial cohort", *Scientific Reports*, Vol. 15, no. 1736, 2025.
[8] G.E, Guraksin and I, Kayadibi, "A Hybrid LECNN Architecture: A Computer-Assisted Early Diagnosis System for Lung Cancer Using CT Images", *International Journal of Computational Intelligence* Systems, Vol. 18, no. 35, 2025.
[9] V, Kumar et al., "Unified deep learning models for enhanced lung cancer prediction with ResNet-50–101 and EfficientNet-B3 using DICOM images", *BMC Medical Imaging*, Vol. 24, no. 63, 2024.
[10] H.F, Al-Yasriy et al., "Diagnosis of Lung Cancer Based on CT Scans Using CNN", *IOP Conference Series: Materials Science and Engineering*, Vol. 928, no. 2, pp. 022035, 2020.
[11] C, Yan and N, Razmjooy, "Optimal lung cancer detection based on CNN optimized and improved Snake optimization algorithm", *Biomedical Signal Processing and Control*, Vol. 86, Part C, pp. 105319, 2023.
[12] H, Al-Yasriy and M, Al-Huseiny, "The IQ-OTH/NCCD lung cancer dataset", Mendeley Data, V4, doi: 10.17632/bhmdr45bh2.4, 2023.
[13] E, Shweikeh et al., "A deep learning model to enhance lung cancer detection using 'Dual-Branch' model classification approach", *PloS one*, Vol. 21, no. 1, e0339404, 2026.
[14] S.G, Armato et al., "The Lung Image Database Consortium (LIDC) and Image Database Resource Initiative (IDRI): a completed reference database of lung nodules on CT scans", *Medical Physics*, Vol. 38, no. 2, pp. 915-931, 2011.

[15] F, Siddiqui et al., "Lung Cancer", *StatPearls*, 2023, https://www.ncbi.nlm.nih.gov/books/NBK482357/
[16] B, Smolarz et al., "Lung Cancer—Epidemiology, Pathogenesis, Treatment and Molecular Aspect (Review of Literature)", *Int. J. Mol. Sci.*, Vol. 26, no. 5, pp. 2049, 2025.
[17] R, Lugano et al., "Tumor angiogenesis: causes, consequences, challenges and opportunities", *Cellular and Molecular Life Sciences*, Vol. 77, no. 9, pp. 1745-1770, 2019.
[18] O, Lababede and M.A, Meziane, "The Eighth Edition of TNM Staging of Lung Cancer: Reference Chart and Diagrams", *The Oncologist*, Vol. 23, no. 7, pp. 844-848, 2018.
[19] C, Beigelman-Aubry et al., "CT imaging in pre-therapeutic assessment of lung cancer", *Diagnostic and Interventional Imaging*, Vol. 97, no. 10, pp. 972-989, 2016.
[20] H.F, Kareem et al., "Evaluation of SVM performance in the detection of lung cancer in marked CT scan dataset", *Indonesian Journal of Electrical Engineering and Computer Science,* Vol. 21, no. 3, pp. 1731-1738, 2021.
[21] H.F, Al-Yasriy et al., The IQ-OTH/NCCD lung cancer dataset [Data set], Kaggle, https://doi.org/10.34740/KAGGLE/DS/672399, 2020.
[22] J, Pearl, "Causal inference in statistics: An overview", *Statistics Surveys*, Vol. 3, pp. 96-146, 2009.
[23] J, Pearl, "Comment: Graphical models, causality and intervention", *Statistical Science*, Vol. 8, no. 3, pp. 266-269, 1993.
[24] J, Pearl, *Causality*, Cambridge university press, 2009.
[25] I, Ng, S, Zhu, Z, Chen and Z, Fang, "A graph autoencoder approach to causal structure learning", *NeurIPS 2019 Workshop*, pp. 1-8, 2019.
[26] X. Zheng et al., "Dags with no tears: Continuous optimization for structure learning", *Advances in neural information processing systems*, arXiv:1803.01422v2, pp. 1-22, 2018.
[27] Y. Yu et al., "DAG-GNN: DAG Structure Learning with Graph Neural Networks", arXiv:1904.10098v1, pp. 1-12, 2019.
[28] K, Bello et al., "DAGMA: Learning DAGs via M-matrices and a Log-Determinant Acyclicity Characterization", arXiv:2209.08037, pp. 1-28, 2022.
[29] D.P, Kingma and M, Welling, "Auto-Encoding Variational Bayes", arXiv:1312.6114v11, pp. 1-14, 2022.
[30] G.J, Szekely et al., "Measuring and testing dependence by correlation of distances", *Ann. Statist.*, Vol. 35, no. 6, pp. 2769-2794, 2007.
[31] B.E, Monroy-Castillo et al., "Improved distance correlation estimation", *Applied Intelligence*, Vol. 55, no. 263, 2025.
[32] Y, Wu et al., "Shannon Entropy based Randomness Measurement and Test for Image Encryption", https://doi.org/10.48550/arXiv.1103.5520, pp. 1-23, 2011.
[33] P, Ramachandran et al., "Searching for Activation Functions", arXiv:1710.05941v2, pp. 1-13, 2017.
[34] A, Chiarenza et al., "Chest imaging using signs, symbols, and naturalistic images: a practical guide for radiologists and non-radiologists", *Insights into Imaging*, Vol. 10, pp. 114, 2019.
[35] K.A, Miles, "Tumour angiogenesis and its relation to contrast enhancement on computed tomography: a review", *European Journal of Radiology*, Vol. 30, no. 3, pp. 198-205, 1999.
[36] M, Yang et al., "CausalVAE: Disentangled Representation Learning via Neural Structural Causal Models", *CVPR 2021*, pp. 9593-9602, 2021.